\documentclass{article}

\usepackage{PRIMEarxiv}

\usepackage{hyperref}
\usepackage{url}
\usepackage{booktabs} 
\usepackage[utf8]{inputenc} 
\usepackage[T1]{fontenc}    
\usepackage{microtype}
\usepackage{fancyhdr}
\usepackage[style=apa]{biblatex}
\usepackage{authblk}
\usepackage{graphicx} 

\input{hyperref-enhanced.tex}

\title{KGATE : a Knowledge Graph Embedding Training Environment
}

\author[1,2]{Benjamin Loire}
\author[1]{Galadriel Brière}
\author[1]{Célia Brahimi}
\author[3]{Antoine Toffano}
\author[1,4]{Anaïs Baudot}
\affil[1]{Aix Marseille Univ., INSERM, Marseille Medical Genetics, Systems Biomedicine Team, Marseille, France}
\affil[2]{Neurology Therapeutic Area, R\&D Servier Paris-Saclay Institut, Gif-sur-Yvette, France}
\affil[3]{LIRMM, Univ. Montpellier, CNRS, Montpellier, France}
\affil[4]{CNRS, Marseille, France}

\begin{document}
\maketitle

\begin{abstract}
Knowledge graph embedding (KGE) models encode the entities and relations of a knowledge graph into a low-dimensional latent space, enabling tasks such as classification or link prediction. Most KGE models follow an autoencoder architecture, in which an encoder projects the knowledge graph into the latent space and a decoder reconstruct it. Combining both encoder and decoder components is increasingly needed, yet existing libraries rarely support complete autoencoders, are often unmaintained, rely on undocumented default hyperparameters, and produce results that cannot be compared across libraries. Here we present KGATE (Knowledge Graph Autoencoder Training Environment), a modular Python library built on PyTorch Geometric and TorchKGE. KGATE lets users assemble initializers, encoders, decoders, losses, negative samplers, and evaluation metrics as building blocks, or plug in their own block. KGATE includes a preprocessing procedure that controls data leakage, a builtin training pipeline, and reproducibility by design. Benchmarks against six existing KGE libraries show that KGATE training time is comparable with the fastest libraries while offering a broader set of features.

Availability: https://github.com/BAUDOTlab/KGATE
\end{abstract}

\keywords{Knowledge Graph Embedding \and Library \and Graph Representation Learning \and Reproductibility}

\section{Introduction}
A Knowledge Graph (KG) is a data structure encoding entities as nodes and their interactions as edges, where the interaction type is explicitly stated. Formally, a KG is described as a set of triplets (head, relation, tail), where head and tail are the origin and destination nodes, and relation is the type of edge connecting them. This structured representation makes KGs particularly suited to represent heterogeneous domain knowledge. For example, biomedical knowledge graphs routinely encode tens of thousands of associations between genes, diseases, and drugs.

Given the scale of many KGs, their manipulation and interpretation is often computationally intractable. Knowledge Graph Embedding (KGE) addresses this by encoding entities and relations into a continuous, low-dimensional latent space. Once trained, a KGE model can be used to perform downstream tasks such as node or edge classification, or link prediction. The classes may correspond to any class label provided by the user, including multi-class biological labels or simply presence/absence in the KG. Link prediction, in contrast, identifies the most likely entity or relation to complete a triplet when given two of the three parts.

The most common architecture for a KGE model is the autoencoder: the encoder projects the graph nodes and edges into a latent space, and the decoder learns how to reconstruct the original graph in a self-supervised training loop. Encoders learn an embedding function that aggregates, for each node, information from its neighbors. This makes them inductive: the learned function applies to nodes unseen during training. Decoders score each triplet with a scoring function. Since each scoring function captures some topologies better than others, decoder performance varies with the relation of interest. Their main limitation is transductivity: they can only score triplets whose nodes were seen during training.

However, most KGE papers use only a decoder with a random initialization for the latent space and no encoder \parencite{boschin_torchkge_2020, lin_learning_2015, bordesTranslatingEmbeddingsModeling2013}. On the other hand, papers introducing novel encoder architectures often use only DistMult as the decoder \parencite{brody_how_2021, hamilton_inductive_2017, gangemi_modeling_2018}. The rise in multimodal and ever-expanding KG datasets requires the inductive abilities of an encoder combined with the decoder specificity. Thus, modern KGE tasks call for an architecture able to implement a robust combination of encoders and decoders.
  
Several libraries including Pytorch Geometric \parencite{fey_pyg_2025}, TorchKGE \parencite{boschin_torchkge_2020}, Ampligraph \parencite{costabello_accentureampligraph_2024}, LibKGE \parencite{broscheit_libkge_2020}, DGL-KE \parencite{zheng_dgl-ke_2020} and PyKEEN \parencite{ali_pykeen_2020} provide frameworks for KGE model training, with implementations of multiple KGE algorithms. Yet, except for PyKEEN and PyTorch Geometric, these libraries are mostly unmaintained. In addition, they are not interoperable, and the results obtained with one library cannot be compared with the results obtained with another (Supplementary materials section 1). Most of these libraries also propagate hardcoded and undocumented default hyperparameters, which might not be the optimal hyperparameters \parencite{prieto_default_2026}. 

In light of the currently limited KGE ecosystem, we propose here a new library, KGATE, for Knowledge Graph Autoencoder Training Environment. KGATE implements complete autoencoders while enforcing best practices and remaining accessible to newcomers. KGATE is built as a set of building blocks to be modular and adapt to any KGE application.

\section{Implementation}
\label{sec:implem}
KGATE, for Knowledge Graph Autoencoder Training Environment, is a modern KGE library built upon Pytorch Geometric \parencite{fey_pyg_2025} and TorchKGE \parencite{boschin_torchkge_2020} while extending the capabilities of both libraries. KGATE enables fast prototyping and offers an extensive KGE API for advanced users. Its main characteristics are : 
\begin{itemize}
    \item A modular autoencoder interface that allows users to assemble building blocks (figure 1) or plug their own code with minimal friction;
    \item A preprocessing procedure that addresses data leakage biases identified in \textcite{briere_benchmarking_2026};
    \item An off-the-shelf training pipeline;
    \item Extensive documentation for new KGE users;
    \item Fully exposed API to build new models;
    \item Reproducibility by design;
    \item Cross-library comparison of models;
\end{itemize}

\begin{figure}[h!]
    \centering
    \includegraphics[width=1\linewidth]{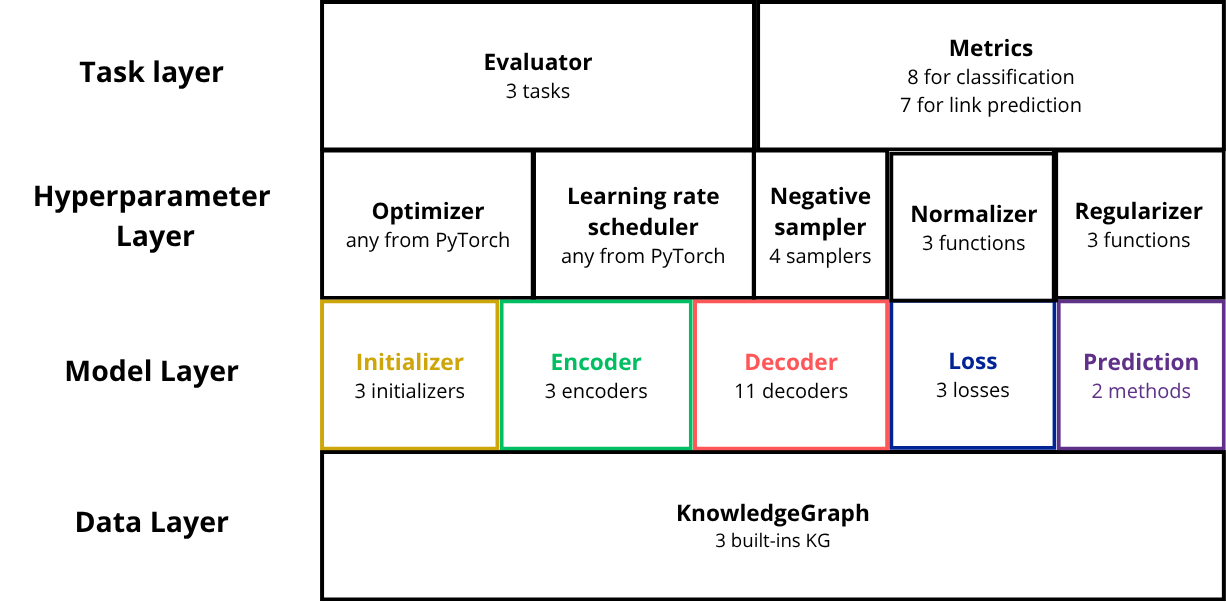}
    \caption{KGATE building blocks. The numbers given for each building block are subject to change as the library evolves. The modular architecture allows for additional blocks to be included by users or  future updates.}
    \label{fig:structure}
\end{figure}

\subsection{Data Layer}

In KGATE, all the data is contained in the KnowledgeGraph class which provides tensor indexing for all triplets, embeddings, mapping dictionaries, split masks, and a range of basic utility for simple graph exploration. As all the data is held in the KnowledgeGraph object, every other building block acts as a transformation function applied to this centralized data. KGATE comes with multiple builtin datasets, such as the standard KGs used in benchmark FB15k-237 \parencite{toutanova_observed_2015} and WN18RR \parencite{dettmers_convolutional_2018}, as well as PrimeKG \parencite{chandak_building_2023}, a biomedical knowledge graph.

KGATE has an extensive data preprocessing toolbox, mainly focused on controlling evaluation biases known as data leakage. Data leakage is described in machine learning as the inclusion in the training dataset of information that should be exclusive to the test dataset, resulting in skewed and inflated performances that do not reflect the real capabilities of a model. While this bias can have a major impact on the relevance of a model in real-life applications, it has been poorly explored in KGE. This issue was first identified in \textcite{akrami_realistic_2020} and explored more in depth in \textcite{briere_benchmarking_2026} (Supplementary materials section 2).

\subsection{Model Layer}
The model layer has four different blocks. First, the initializer manages the initial feature vectors for the nodes and edges. Currently, KGATE provides three types of initialization:
A random initialization using a Xavier uniform function for each node and edge type.
Initialization with user-supplied features such as text embeddings.
Using a Node2Vec \parencite{grover_node2vec_2016} model to generate topology-aware initial feature vectors.

During training, the encoder learns to project these features into a low-dimensional latent space using the graph structure. Encoders are graph neural networks able to aggregate information from heterogeneous features. KGATE implements 3 encoders from PyTorch Geometric: GATv2 \parencite{brody_how_2021}, RGCN \parencite{gangemi_modeling_2018} and GraphSAGE \parencite{hamilton_inductive_2017}. It is also possible to use no encoder at all, in which case the initial feature vector output of the initializer is directly forwarded as the decoder’s input.

After the encoder’s forward pass, the decoder attempts to reconstruct the original graph from the latent space by computing a score for each triplet according to its own scoring function. KGATE implements 11 decoders from the three main families of KGE models:
\begin{itemize}
    \item Translational models: also called geometric KGE models, they score the triplets so that the translation of the head node according to the relation is close to the tail node. KGATE implements the following translational models: TransE \parencite{bordesTranslatingEmbeddingsModeling2013}, TransH \parencite{wang_knowledge_2014}, TransR \parencite{lin_learning_2015}, TransD \parencite{ji_knowledge_2015}, TorusE \parencite{ebisu_toruse_2018} and RotatE \parencite{sun_rotate_2018}.
    \item Bilinear models: they use matrix factorization as the score function. KGATE implements the following bilinear models: RESCAL \parencite{nickel_three-way_2011}, DistMult \parencite{yang_embedding_2014} and ComplEx \parencite{trouillon_complex_2016}.
    \item Convolutional models: they are deep learning algorithms using convolutional neural networks as the score function. KGATE implements ConvKB \parencite{nguyen_novel_2018} and ConvE \parencite{dettmers_convolutional_2018}.
\end{itemize}
By default, each decoder’s configuration is identical to its original paper’s implementation, but users can freely override these defaults to explore the hyperparameter space.

In addition, KGATE provides multiple loss functions (margin ranking, binary cross-entropy and logistic losses), embedding regularizers and normalizers. Finally, KGATE supports two prediction methods: the regular rank-based prediction method or the set retrieval technique introduced in \textcite{li_sphere_2024} a, which returns a set of true triplets. 

\subsection{Hyperparameter Layer}
This layer holds all the configurable aspects of the training. KGATE can use any optimizer and learning rate scheduler provided by PyTorch. This layer encapsulates all training utilities, such as the embedding normalizer and the regularizer which are respectively applied before and after every gradient step, and the negative sampler. Indeed, KGE training relies on the generation of false (negative) triplets that the decoder should score lower than true triplets. KGATE provides four different negative samplers: Positional, Bernoulli, Uniform and Mixed. The choice of negative triplet generation method and the complexity of these false triplets is an essential hyperparameter. 

\subsection{Task Layer}
A KGATE model is trained to perform one of three tasks: Node/Edge Classification, Triplet Classification or Link Prediction. Depending on the task, a range of metrics are available to evaluate the model’s performances. Link prediction metrics are: mean reciprocal rank (MRR), mean rank (MR), hit@k, median rank, MRR@k, score gap and relative rank; Classification tasks metrics consist of accuracy, precision, recall, specificity, F1, false negative and false positive rates. Helper objects allow for an easy extension of available metrics.

\subsection{Modularity and Extensibility}
While KGATE was created to lower the expertise threshold required to perform KGE analyses, it is also a suitable tool for experts looking for a reproducible autoencoder framework. Each block of a KGATE workflow can be customized or replaced entirely, and additional building blocks can be inserted. For instance, new tasks and metrics can be added directly to the main KGATE structure. Moreover, KG embeddings can be exported to be used externally.

\section{Benchmarking KGATE in the KGE library ecosystem}
\label{sec:benchmark}
We benchmarked the technical performance of KGATE against six other KGE libraries: PyTorch Geometric \parencite{fey_pyg_2025}, PyKEEN \parencite{ali_pykeen_2020}, TorchKGE \parencite{boschin_torchkge_2020}, Ampligraph \parencite{costabello_accentureampligraph_2024}, DGL-KE \parencite{zheng_dgl-ke_2020} and LibKGE \parencite{broscheit_libkge_2020}, on two standard KG datasets, FB15k-237 and WN18RR. We compared data loading, mean training epoch, evaluation durations, and evaluation metrics  (supplementary materials section 3). On average, KGATE’s speed is comparable to the best-performing libraries and is orders of magnitude faster than PyKEEN, which is the most similar library in terms of utilities.

\section{Conclusion}
KGATE provides utilities absent from common KGE libraries, including a complete autoencoder architecture and a built-in data leakage control procedure, gathered in a fully modular structure. KGATE is released in version 1.0 with different extensions planned to integrate explainability and temporal KGE modules.

\section{Code Availability}
The code of KGATE is available at \href{https://github.com/BAUDOTlab/KGATE}{https://github.com/BAUDOTlab/KGATE}

The code of the benchmark is available at \href{https://github.com/BAUDOTlab/KGATE_benchmark}{https://github.com/BAUDOTlab/KGATE\_benchmark}

\section*{Acknowledgments}
This work was supported by members of Servier R\&D and in particular Philippe Moingeon, Nicolas Lévy and Céline Lefebvre at Servier.

\section{Fundings}
This work was primarily funded by Servier.
\newline
The aims of this study contribute to the ERDERA project, which has received funding from the European Union’s Horizon Europe research and innovation program under grant agreement N°101156595.
\newline
The project leading to this publication has received funding from the Excellence Initiative of Aix-Marseille Université - A*Midex, a French “Investissements d’Avenir programme” AMX-21-IET-017.
\printbibliography

\end{document}


\maketitle

\makeatletter
\renewcommand{\thetable}{S\arabic{table}}
\renewcommand\fnum@table{Supplementary Table~\thetable}
\renewcommand{\thefigure}{S\arabic{figure}}
\renewcommand\fnum@figure{Supplementary Figure~\thefigure}
\renewcommand{\theequation}{S\arabic{equation}}
\makeatother

\renewcommand{\thesection}{Supplementary Section S\arabic{section}: }
\renewcommand{\thesubsection}{\thesection.\arabic{subsection}}
\renewcommand{\thesubsubsection}{\thesubsection.\arabic{subsubsection}}

\renewcommand{\arraystretch}{1.5}

\section{Main differences between KGE libraries}
\label{supp1}
Knowledge Graph Embedding (KGE) models, just like any machine learning models, rely on hyperparameters to guide their training. In order to train a reproducible model, these hyperparameters must be kept identical between training sessions. In theory, identical model hyperparameters should output the same results in the same conditions, regardless of the library running the code. However, the original publications of KGE models are sometimes ambiguous about these hyperparameters, or do not provide the source code. In turn, developers trying to implement these models are forced to interpret the author’s intent and make arbitrary decisions.

In addition, some KGE libraries present differences in model implementation. For example, in the original DistMult publication (Yang et al. 2014), the objective formula provided in the manuscript does not indicate embedding normalization, but reads “The entity vectors are renormalized to have unit length after each gradient step [...]. For the relation parameters, we used standard L2 regularization.” and the code is not provided by the authors. As a consequence, the DisMult model has been implemented differently in different libraries:
\begin{itemize}
    \item TorchKGE performs L2 normalization on the entity vectors at the beginning of each epoch.
    \item PyTorch Geometric, Ampligraph, and DGL-KE, do not normalize nor regularize anything.
    \item PyKEEN performs L1 normalization on entity embeddings at the beginning of each epoch and regularizes the edge parameters after each gradient step.
\end{itemize}

These design choices regarding hyperparameters and implementation are often hardcoded, which makes it impossible for the end user to alter without editing the library’s code, and breaks cross-library reproducibility. The table below illustrates these variations on two standard KGE datasets with the following hyperparameters:

\begin{itemize}
\item Initialization: Random
\item Encoder: None
\item Decoder: TransE
\item Embedding dimension: 256
\item Loss: Margin ranking loss
\item Margin: 0.5
\item Negative Sampler (when tunable): Bernoulli
\item Number of negatives per positive triplet: 5
\item Batch size: 4096
\item Epochs: 100
\item Learning Rate: 0.001
\item Optimizer: Adam
\item Random seed: 42    
\end{itemize}

\begin{figure}[h!]
    \centering
    \includegraphics[width=1\linewidth]{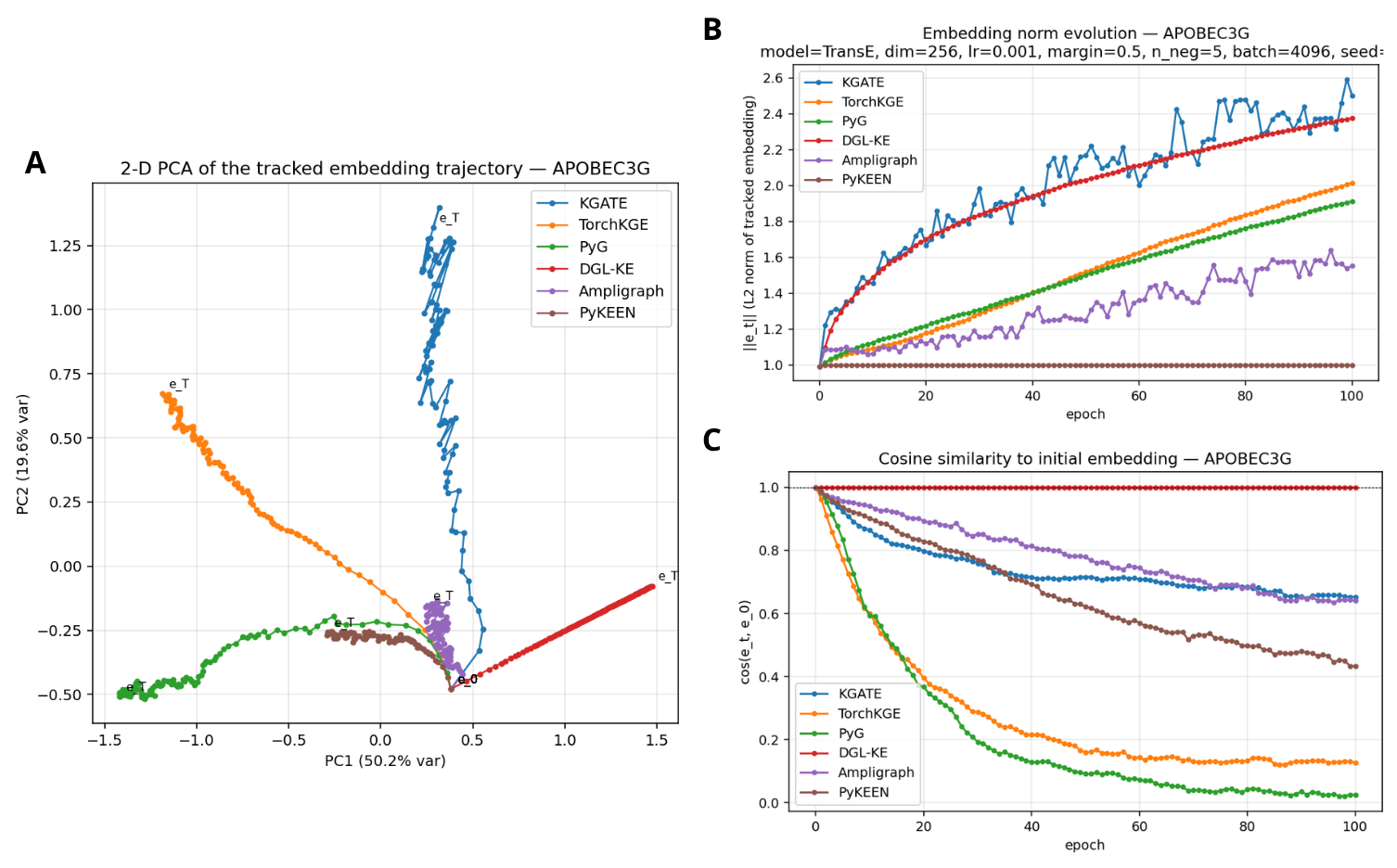}
    \caption{Evolution of the embeddings of one gene node across 100 epochs using the training loops of five different libraries on the PrimeKG dataset. The embeddings were initialized once before being fed to each library as initial features to the nodes. Each run was performed on a single-threaded CPU to ensure determinism. A) Evolution of the two first components of a PCA performed on the embeddings across each epoch; B) Evolution of the L2 normalization of the embedding; C) Cosine similarity of the embedding at each epoch compared to the initial embedding.}
    \label{fig:embedding_evolution}
\end{figure}

\begin{table}[h!]
    \centering
    \begin{tabular}{|c|c|c|c|c|c|c|}
    \hline
        Library & KGATE & TorchKGE & PyTorch Geometric & DGL-KE & Ampligraph & PyKEEN\\
    \hline
        KGATE & / & 0.290 & 0.169 & 0.070 & 0.107 & 0.284\\
    \hline
        TorchKGE & 0.290 & / & 0.260 & 0.058 & 0.077 & 0.197\\
    \hline
        PyTorch Geometric & 0.0169 & 0.260 & / & 0.053 & 0.058 & 0.114\\
    \hline
        DGL-KE & 0.070 & 0.058 & 0.053 & / & 0.053 & 0.064\\
    \hline
        Ampligraph & 0.107 & 0.077 & 0.058 & 0.053 & / & 0.070\\
    \hline
        PyKEEN & 0.284 & 0.197 & 0.114 & 0.064 & 0.070 & /\\
    \hline
    \addlinespace
    \end{tabular}
    \caption{Mean Jaccard index of the similarity of the top 10 predictions for 10 different gene nodes sampled at random using the link prediction query “(gene, protein-protein, ?)”. Predictions are made once 100 epochs have elapsed, using the same model and the same hyperparameters when it is possible to tune them.}
    \label{tab:jaccard_table}
\end{table}

While all libraries are deterministic given the same seed and hyperparameters, the differences between their implementations and hyperparameters (e.g., normalization, loss and scoring functions) lead to a different output, which affects the reproducibility between libraries. These differences lead to very different MRRs and different predictions. This means that results may be driven more by the choice of the library rather than by model capabilities. From a reproducibility standpoint, this implies that model comparisons and benchmarks must be done using the exact same library. When that is not the case, a difference in results cannot be directly attributed to the underlying data or methodological approach.

To enable a fair comparison between libraries, KGATE provides cross-library comparison through configuration presets able to reproduce the results given by a different library.

\section{Details on the Data Leakage control procedure}
\label{supp2}

When the train/test split of a knowledge graph is not properly controlled, the training set may contain triplets that are redundant with test triplets. Such redundancy biases the evaluation by artificially inflating the measured performance of KGE models, an issue known as data leakage (\textcite{akrami_realistic_2020, briere_benchmarking_2026}). To address it, KGATE implements a data leakage control procedure that detects redundant relations in the KG and removes the triplets inducing data leakage from the training set.

Based on prior studies by Akrami et al. and Brière et al., the data leakage procedure implemented in KGATE is able to detect three types of relations that may cause data leakage issues in a KG: near-duplicate relations, near-reverse-duplicate relations, and Cartesian product relations. 

In a biomedical KG, two relations types r1 and r2 may convey similar meanings, causing a certain proportion of their (head, tail) pairs to overlap. For example, “(protein)-binds to-(protein)” and “(protein)-interacts with-(protein)” are expected to link many of the same protein pairs, since binding is a form of interaction. Given two relations r1 and r2, when the overlap between their (head, tail) pairs in the untreated KG exceeds a user-defined threshold (often set to 80\%, the default value in KGATE),  they are considered as near-duplicate relations. To account for asymmetric redundancies, the overlap is computed in both directions (relative to r1 and relative to r2), and the user can specify a distinct threshold for each direction. 

Similarly, reverse relations, such as “(drug)-inhibits-(protein)” and “(protein)-is inhibited by-(drug)” can be considered as near-reverse-duplicates as they convey reverse semantic meanings. Near-reverse-duplicate relations are detected with the same procedure to near-duplicate relations, but comparing the (head, tail) pairs of r1 with the (tail, head) pairs of r2. 

The third type of redundant relations that can be detected with KGATE’s data leakage procedure are Cartesian product relations. Let us consider the set of unique heads and the set of unique tails connected by a relation r. The maximum number of distinct triplets that r could form between its set of heads and its set of tails is the product of the sizes of these two sets, corresponding to every head being connected to every tail. A relation r is considered a Cartesian product relation when the number of triplets it actually forms in the full KG exceeds a user-defined threshold of this maximum. In a biomedical KG, an example of such Cartesian product relation could be “(gene)-expressed in-(cell type)” for ubiquitously expressed genes (i.e. housekeeping genes).

After detecting near-duplicate relations, near-reverse-duplicate relations, and Cartesian product relations in the KG, KGATE performs the train/test split and proceeds to remove any leaked triplet from the training set.  Specifically, for near-duplicate and near-reverse-duplicate relations, for any triplet placed in the validation or test set, its redundant counterparts are excluded from the training set. For Cartesian product relations, KGATE ensures that all triplets involving a given head (or tail) with the Cartesian relation r are placed in the same split.

While most KGE models handle exclusively directed relations, biomedical KGs often contain undirected relations (e.g., protein-protein interactions). KGATE can convert these undirected relations into directed ones while ensuring that this conversion does not introduce near-reverse-duplicate relations and data leakage in consequence. For any undirected relation r, KGATE creates a reverse counterpart r\_rev and adds, for every triple (a, r, b), the corresponding triple (b, r\_rev, a). Then, for every test triple (a, r, b), the reverse triple (b, r\_rev, a) is removed from the training set, and vice versa. Importantly, all triplets are kept in KGATE's ground-truth. The ground-truth triplets are removed in the filtered evaluation, so that models are not penalized for ranking highly triplets that represent true facts but were removed from the training graph. For instance, since protein-protein interactions are undirected, a test triple (a, ppi, b) implies that (b, ppi, a), (a, ppi\_rev, b) and (b, ppi\_rev, a) are also true, and all of them are kept in the ground truth, even though they are removed from the training set.

\newpage
\section{Benchmarking KGATE and other libraries}
\label{supp3}

\begin{table}[h!]
    \centering
    \setlength{\tabcolsep}{10pt}
    \begin{tabular}{ccccccc}
    \hline
        Library & Encoders & Decoders &\thead{KGE \\Autoencoder} & \thead{Negative \\samplers} & \thead{Data Leakage\\Control} & \thead{Actively\\Maintained}\\
    \hline
        \textbf{KGATE (ours)} & 3 & 10 & Yes & \textbf{4} & \textbf{Advanced} & Yes\\
        PyTorch Geometric & \textbf{21} & 4 & No & 1 & Basic & Yes\\
        TorchKGE & 0 & 11 & No & 3 & No & No\\
        PyKEEN & 3 & \textbf{34} & One & 3 & Basic & Yes\\
        LibKGE & 0 & 11 & No & 2 & No & No\\
        Ampligraph & 0 & 4 & No & 1 & No & No\\
    \hline
    \addlinespace
    \end{tabular}
    \caption{Comparisons of KGATE and other KGE libraries utilities. The advanced data leakage of KGATE corresponds to the full procedure described in Brière et al, while Basic data leakage is a simple control of full duplicate edges. LibKGE appears to provide the necessary structure for an autoencoder; however no encoder is implemented. PyKEEN does implement a kind of autoencoder, with the implementation of specific message passing encoders like RGCN that take into account the different edge types. However, in PyKEEN, the generic message passing intended for user-driven customization treats the dataset as a homogeneous graph instead of a KG, failing to represent the complexity of the data. }
    \label{tab:feature_table}
\end{table}

\begin{table}[h!]
    \centering
    \label{tab:fb15k237_table}
    \textbf{FB15k-237}
    \begin{tabular}{lcccc}
        \hline
        Library & Data loading (s) & Mean epoch (ms) & Test (s) & MRR \\
        \hline
        KGATE (ours) & 3.09 & 0.92 & 4.57 & 0.2340 \\
        TorchKGE & 3.92 & 0.21 & 5.51 & 0.2419 \\
        PyTorch Geometric & 0.10 & 0.21 & 9.19 & 0.2381 \\
        PyKEEN & NA & 7.11 & 3.89 & 0.0791 \\
        Ampligraph & 1.73 & 0.22 & 44.01 & 0.1547 \\
        LibKGE & 0.50 & 0.87 & 7.88 & 0.0592 \\
        DGL-KE & 0.49 & 0.46 & 5.29 & 0.0870 \\
        \hline
    \addlinespace
    \end{tabular}
    \textbf{WN18RR}
    \begin{tabular}{lcccc}
        \hline
        KGE Library & Data loading (s) & Mean epoch (ms) & Test (s) & MRR \\
        \hline
        KGATE (ours) & 1.01 & 0.44 & 2.93 & 0.0143 \\
        TorchKGE & 1.30 & 0.11 & 2.29 & 0.0060 \\
        PyTorch Geometric & 0.05 & 0.18 & 0.14 & 0.0001 \\
        PyKEEN & NA & 2.41 & 2.15 &  \\
        Ampligraph & 0.18 & 0.15 & 6.54 & 0.0967 \\
        LibKGE & 0.46 & 0.42 & 4.15 & 0.0018 \\
        DGL-KE & NA & NA & NA & NA \\
        \hline
    \addlinespace
    \end{tabular}
    \caption{Training duration and reported metric for different libraries trained with the TransE decoder on the FB15k-237 and WN18RR KG datasets using the same set of hyperparameters. For KGATE, the data loading duration was measured without using KGATE’s preprocessing pipeline. PyKEEN’s data loading duration is set to NA because it was not possible to separate it from the training step, however it was negligible compared to the duration of a training epoch. We did not manage to run DGL-KE on the WN18RR dataset.}
\end{table}

KGATE has a similar or slightly longer training time than other libraries, but is 4 to 5 times faster than PyKEEN, which is the only library that provides a similar range of utilities. Except for PyKEEN, the duration differences in overall training time over a full training of 100 epochs are negligible.